\documentclass[10pt,twocolumn,letterpaper]{article}
\usepackage{colortbl} 
\usepackage{xcolor}   
\usepackage{cvpr}      
 \usepackage[ruled,vlined]{algorithm2e}
\usepackage{multirow}

\usepackage{CJKutf8}
\newcommand{\ourmethod}{AdaScope}
\definecolor{cvprblue}{rgb}{0.21,0.49,0.74}
\usepackage[pagebackref,breaklinks,colorlinks,allcolors=cvprblue]{hyperref}
\usepackage{amsthm}

\def\paperID{----} 
\def\confName{CVPR}
\def\confYear{2026}
\newtheorem{theorem}{Theorem}
\newtheorem{lemma}{Lemma}

\title{Do Less, Achieve More: Do We Need Every-Step Optimization for RL Fine-tuning of Diffusion Models?}

\author{
Renye Yan$^{1}$\thanks{Equal contribution}, Jikang Cheng$^{1*}$, Shikun Sun$^{2}$, Yi Sun$^{2}$, You Wu$^{3}$, Wei Peng$^{4}$, Zongwei Wang$^{1}$, \\Ling Liang$^{1\dagger}$, Junliang Xing$^{2}$, Yimao Cai$^{1}$\thanks{Corresponding author} \\[0.2cm]
$^{1}$Peking University, $^{2}$Tsinghua University, $^{3}$Nanjing University,
$^{4}$Stanford University 
}

\begin{document}
\begin{CJK*}{UTF8}{gbsn}

\maketitle
\begin{abstract}

Despite strong image-generation performance, diffusion models’ reconstruction objectives limit alignment with human preferences. RL enables such alignment through explicit rewards. However, most studies apply RL to the full denoising trajectory, making it computationally costly and weakening preference alignment, i.e., doing more but achieving less. We observe that the impact of RL fine-tuning varies significantly across denoising stages. In the early stage, image structures are unstable and distant from the final reward signal. Applying RL at this stage leads to delayed rewards and action–reward mismatching, resulting in high variance and inefficient updates. Conversely, in the later stage, reward gains saturate, and continued training tends to overfit local details, intensifying reward hacking. To tackle these challenges, we propose \ourmethod{}, an RL-enhanced plug-in that improves generation quality while reducing computational cost. Specifically, \ourmethod{} adaptively identifies the optimal intervention timing for RL by perceiving the structural evolution and semantic consistency during denoising, and dynamically terminates training once the denoising converges and reward gains saturate. As a result, it achieves a rare `dual benefit': a reduction in computational costs alongside a significant performance improvement. We offer theoretical grounds for the design of \ourmethod{}. Compared with state-of-the-art methods, \ourmethod{} improves performance by 66\% while cutting computational cost by 59\%.

\end{abstract}

\begin{figure}[t]
    \centering
    \includegraphics[width=1\linewidth]{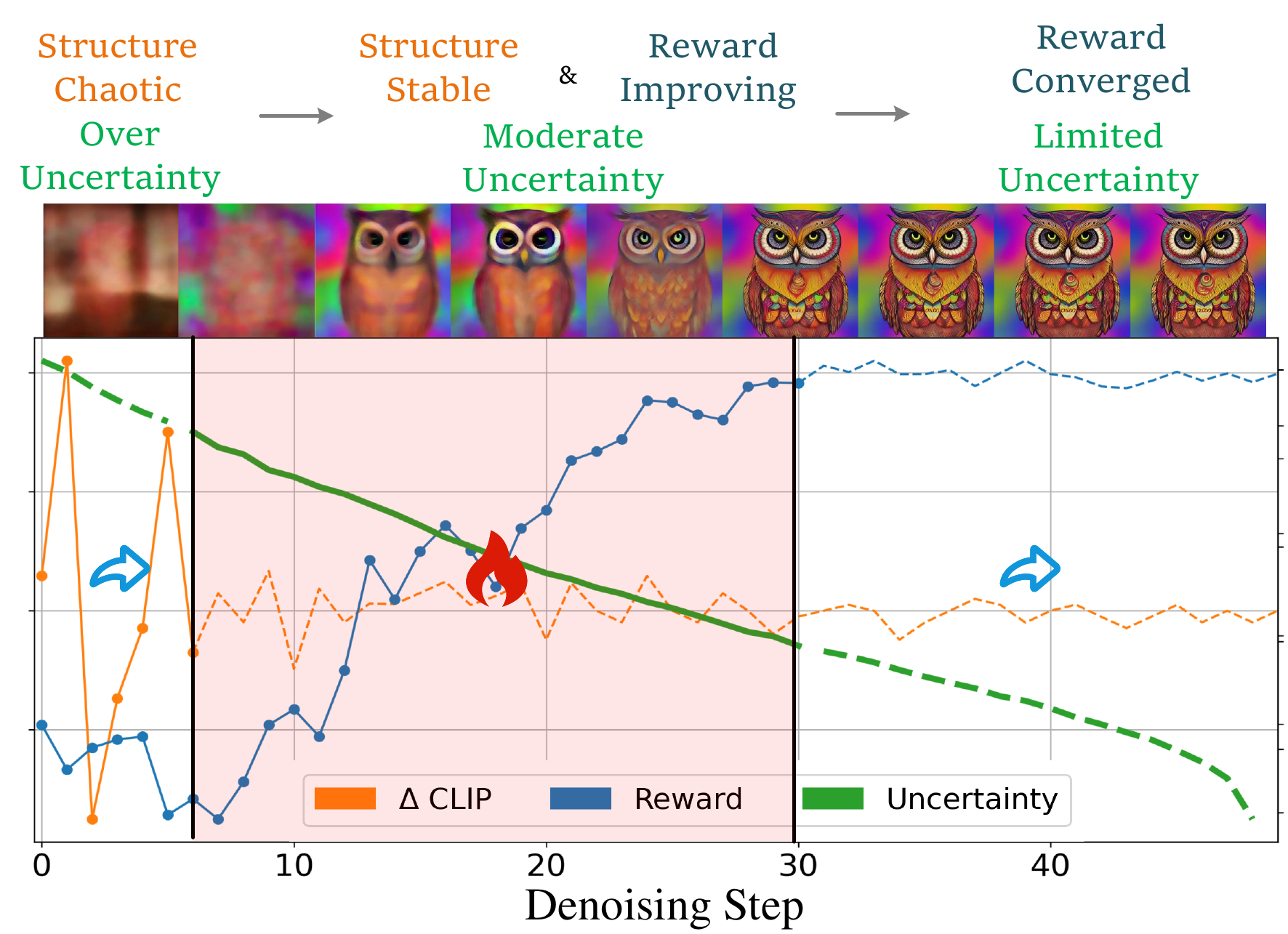}
    \caption{We plot the CLIP variation ($\Delta$ CLIP), Reward Objective, and Uncertainty Score (Based on Lemma~\ref{correlation-lemma}) with aligned denoising steps. Only the red region is optimized, where we leverage $\Delta$ CLIP and Reward to select the adaptive scope of denoising steps for training. It can be observed that the structure is chaotic in the first stage, while the reward converges in the last stage. The selected scope has a stable structure and improving reward, which exactly corresponds to the moderate uncertainty stage.}
    \label{fig:intro}
\end{figure}
\section{Introduction}
\label{sec:intro}

Diffusion models~\cite{sohl2015deep,ho2020denoising,nichol2021glide,ramesh2022hierarchical,rombach2022ldm,saharia2022photorealistic,croitoru2023diffusion,chen2023diffusiondet,kingma2021variational,wei2026team,zou2025mixturegloballocalexperts} have emerged as the core framework for general-purpose image generation, owing to their powerful distribution modeling capability and exceptional generation quality. By progressively denoising to approximate the true distribution, they can faithfully reconstruct complex multimodal structures and have achieved groundbreaking progress in fields such as vision, speech, and 3D generation~\cite{ho2022imagen,ho2022video,xu2023dream3d,zhou20213d}. However, the optimization objective of such models primarily focuses on data reconstruction, lacking task-specific or goal-oriented constraints, such as aesthetic preferences~\cite{aesthetic} or user-specific intents~\cite{yang2024using}. As a result, although these models possess strong generative potential, they struggle to achieve fine-grained alignment with specific task. 

Recent studies have attempted to incorporate reinforcement learning (RL)~\cite{schulman2017proximal,reflectivepo,zhao2022alphaholdem,yan2024adazero} into the diffusion model fine-tuning, enabling reward signals to guide the generation process toward goal-oriented optimization directly ~\cite{black2023training,fan2024reinforcement,yang2024using}. RL methods~\cite{black2023training,oertell2025efficient,fan2024reinforcement,yang2024using,wallace2024diffusion,yan2025entropy} extend the optimization objective from data reconstruction to preference consistency, thereby achieving task-level alignment. However, the introduction of RL inevitably brings the issue of sparse feedback. Since rewards are only available after the final image is generated, the early denoising stage lacks effective guidance signals, making policy optimization difficult. To address this, mainstream approaches typically adopt a straightforward compromise: backpropagating the final reward to all denoising steps to ensure RL functions properly~\cite{fan2023dpok,black2023training,franceschelli2024reinforcement,chen2024overview,uehara2024understanding}. In addition, the large computational overhead introduced by RL fine-tuning is also highly significant~\cite{kim2025test,xie2025dymo}.

Although RL fine-tuning methods have achieved remarkable results, they still face three critical unresolved challenges. (1) Temporal Misalignment Caused by Reward Backfilling:
While backpropagating the final reward to all denoising steps mitigates sparse feedback, it disrupts temporal causality. Early steps receive guidance signals weakly correlated with their actual contributions. Moreover, due to the rapid structural changes in the image during the early stage, RL training tends to have high variance and random gradient directions~\cite{fan2023dpok,black2023training,franceschelli2024reinforcement,chen2024overview,uehara2024understanding}. (2) Aggravated Reward Hacking Due to Neglected Stage Sensitivity:
Existing methods typically align the training termination point with the final denoising step, implicitly assuming equal optimization value across all stages. This assumption overlooks the fact that structural and feature representations become highly stable in the late stage, where denoising has nearly converged. The marginal benefit of RL training at this stage is minimal. Forcing model updates in such low-value intervals increases the risk of amplifying reward hacking~\cite{skalse2022hacking,gao2023scaling,zhang2024confronting,miao2025inform}. (3) Negative Impact and Computational Waste:
The issues described in (1) and (2) not only hurt final generation quality but also lead to significant computational resource waste~\cite{tang2024inference,kim2025test,xie2025dymo,black2023training,fan2023dpok,yang2024using,franceschelli2024reinforcement}.

To address the issues above systematically, we propose \ourmethod{}, an RL plugin that adaptively screens training stages. Unlike the current training paradigm of optimizing full trajectories at equal frequency, \ourmethod{} dynamically identifies the optimal intervention and termination points for RL. Specifically, \ourmethod{} uses semantic maturity as a criterion to adaptively determine the starting point for RL, skipping the chaotic stage in which structural formation is incomplete and noise predominates. This helps alleviate reward-attribution mismatch. Subsequently, by jointly modeling denoising completeness and reward-gain trends, \ourmethod{} automatically stops updates when marginal returns saturate, thereby effectively suppressing low-return training and reward hacking. Ultimately, the optimization process focuses on the key interval where semantic structures are stable and training returns are high. Since \ourmethod{} dynamically filters out denoising stages that contain a large number of inefficient or counterproductive samples, it not only optimizes computational resources but also delivers significant improvements in generation quality. We validate the design rationale of \ourmethod{} by visualizing the correlation between adjacent intermediate denoising steps and conducting corresponding theoretical analyses.

To verify the integrability and effectiveness of \ourmethod{}, we integrated it with 4 different approaches. The results demonstrate that our method significantly outperforms state-of-the-art methods across 6 metrics while reducing computational costs to 59\%. In summary, this work presents the following three key innovations:

\begin{itemize}
    \item We demonstrate the rationality of our method’s design through theoretical derivations of the correlations among intermediate images during the generation process. Furthermore, we provide a visualization analysis of the correlation metric, whose observed results are consistent with the derived conclusions.

    \item In terms of quality, \ourmethod{} automatically filters out early-stage samples with severe reward attribution mismatch, thereby enhancing the stability of policy gradient estimation. Simultaneously, it eliminates late-stage samples that tend to overfit, thereby strengthening generalization and distribution fidelity.

    \item From a computational perspective, \ourmethod{} reduces training steps by adaptively identifying effective RL intervals, thus avoiding ineffective updates in early high-noise and late-saturation stages. This design yields higher returns under the same computation cost.

\end{itemize}

\section{Related Works}

\paragraph{Sparse Rewards:}
\label{sparse_reward_related_work}

In RL fine-tuning for diffusion models, reward signals are typically computed precisely only on the final generated image, resulting in a sparse rewards~\cite{huang2024diffusion,ma2024deepcache,hare2019dealing,wang2020deep,riedmiller2018learning,goecks2019integrating} profile during fine-tuning. 
The diffusion model must rely on distant terminal signals to make updates across the extended denoising process. This temporal disconnection disrupts optimization causality and makes it difficult to accurately assess the contribution of actions in the early stage.

\begin{figure}
    \centering
    \includegraphics[width=0.9\linewidth]{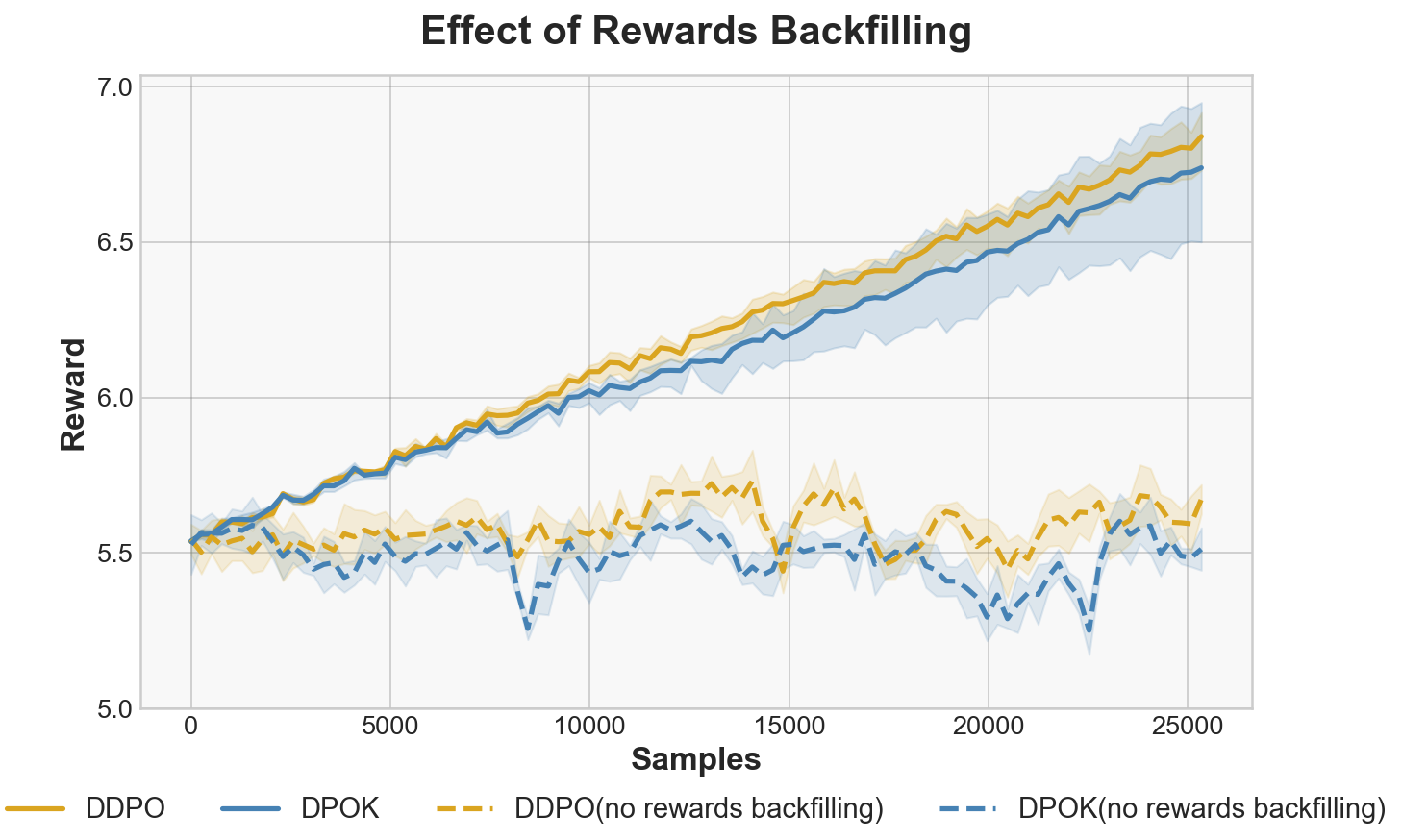}
    \caption{In T2I generation, rewards can only be computed after denoising is fully completed, resulting in extremely sparse feedback that cannot support stable training (see the dashed curves).}
    \label{fig:reward_backing}
\end{figure}

The resulting credit assignment problem leads to increased gradient variance, slower convergence, and significantly reduces the RL fine-tuning efficiency in diffusion models. To maintain trainability under sparse reward conditions, the common practice is to backpropagate the final reward to all denoising steps. While this mechanism partially mitigates the signal deficiency, it substantially increases computational overhead and may introduce unreasonable optimization biases (see Fig.~\ref{fig:reward_backing}).

\paragraph{Reward Hacking:}

Another issue in RL fine-tuning for diffusion models is that the reward signals are typically based on external metrics of the generated images, such as aesthetic scores or semantic consistency ratings. However, when the diffusion model directly optimizes these metrics through RL, `reward hacking' ~\cite{skalse2022defining,eisenstein2023helping,yuan2019novel,hu2025reward}often occurs: to maximize the reward signal, the diffusion model learns strategies that deviate from the true semantic or visual distribution. For example, it may generate images with higher contrast, sharper yet distorted features, thereby causing inflated evaluations from the scoring signals at the expense of realism and diversity. Essentially, this arises because the reward function fails to fully capture the distribution of human preferences, leading to semantic misalignment between the optimized objective and the intended goal. As fine-tuning iterations accumulate, this deviation is continuously amplified, causing the model to overfit to pseudo-optimal solutions of the reward model and resulting in a high-reward but low-quality generation mode. The reward backpropagation mechanism further exacerbates the reward hacking problem. To prevent the model from fooling the reward model during RL, DPOK~\cite{fan2024reinforcement} and its derivative methods attempt to impose KL regularization constraints on the policy, keeping the generated distribution near the original model's manifold. While such approaches partially mitigate reward hacking, they do not explicitly address the elimination of computational overhead. For more detailed Related Works, please refer to Supplementary Material.
\section{Method}
\label{sec:method}

\begin{figure*}
    \centering
    \includegraphics[width=1.0\linewidth]{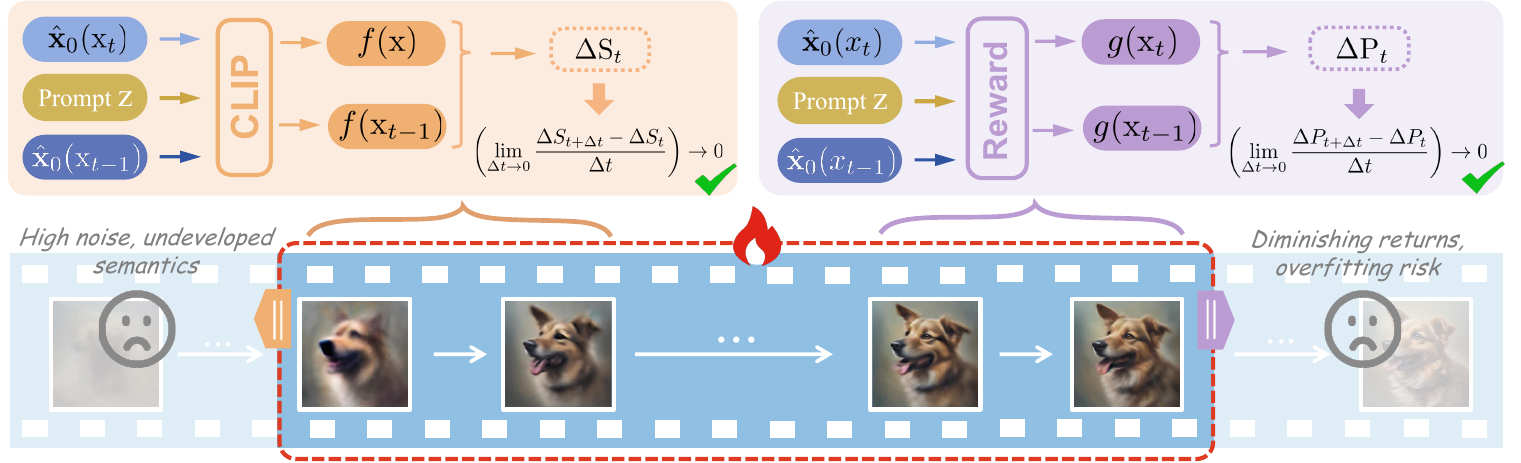}
    \caption{Overall framework of our method}
    \label{fig:fromwork}
\end{figure*}

To achieve the dual objective of significantly reducing computational cost while improving training quality, we propose \ourmethod{}. Specifically, Sec.~\ref{PF} and Sec.~\ref{mdp} introduce the necessary notation and fundamental formulations. In Sec.~\ref{theory_method}, we provide the theoretical basis for the rationality of \ourmethod. Finally, Sec.~\ref{scop_selection} elaborates on the underlying adaptive mechanism of \ourmethod{}.

\subsection{Problem Formulation}
\label{PF}
\paragraph{Diffusion Models.}
A diffusion model defines a data distribution by inverting a fixed 
noising process.  
Starting from a clean data point $\mathbf{x}_0$, the forward 
diffusion process 
$q(\mathbf{x}_t \mid \mathbf{x}_{t-1})$ 
adds gradually increasing Gaussian noise, producing a latent trajectory 
$\mathbf{x}_0 \!\rightarrow\! \cdots \!\rightarrow\! \mathbf{x}_T$, 
where the terminal state $\mathbf{x}_T$ is distributed as 
$p(\mathbf{x}_T)=\mathcal{N}(0,I)$.  
The generative model learns the reverse-time conditionals
\[
p_\theta(\mathbf{x}_{0:T})
= p(\mathbf{x}_T)\prod_{t=1}^T 
p_\theta(\mathbf{x}_{t-1}\mid \mathbf{x}_t),
\]
which are trained to match the true reverse posterior 
$q(\mathbf{x}_{t-1}\mid \mathbf{x}_t,\mathbf{x}_0)$.  
Sampling begins from noise $\mathbf{x}_T$ and repeatedly applies 
the learned reverse transitions to reconstruct a data sample.
Let 
\[
q\left(\mathbf{x}_t \mid \mathbf{x}_{t-1}\right)
:= \mathcal{N}\!\left(\mathbf{x}_t; \sqrt{1 - \beta_t}\,\mathbf{x}_{t-1},\, \beta_t \mathbf{I}\right),
\]
where
\[
\alpha_t := 1 - \beta_t, 
\qquad
\bar{\alpha}_t := \prod_{k=1}^t \alpha_k, 
\quad 
(\bar{\alpha}_0 := 1).
\]
Suppose $\epsilon_\theta(\mathbf{x}_t, t)$ is the learned noise-prediction function.  
Then the corresponding $\hat{\mathbf{x}}_0$ can be obtained as
\begin{equation}
\label{def: pseudo x0}
\hat{\mathbf{x}}_0(\mathbf{x}_t, t)
= \frac{1}{\sqrt{\bar{\alpha}_t}}
\left(
    \mathbf{x}_t
    - \sqrt{1 - \bar{\alpha}_t}\,
      \epsilon_\theta(\mathbf{x}_t, t)
\right).
\end{equation}

\paragraph{Text-to-Image Tasks.}
For text-conditioned generation, a conditioning representation
$\mathbf{z}$, typically extracted from a natural-language prompt, is 
introduced to guide the reverse diffusion process.
The model therefore parameterizes
\[
p_\theta(\mathbf{x}_{0:T}\mid \mathbf{z})
= p(\mathbf{x}_T)\prod_{t=1}^T 
p_\theta(\mathbf{x}_{t-1}\mid \mathbf{x}_t,\mathbf{z}),
\]
so that each denoising step is guided by $\mathbf{z}$.  
This conditioning mechanism steers the reverse trajectory toward samples 
that are semantically consistent with the input text.

\subsection{AdaScope’s MDP Formulation}
\label{mdp}

We formulate the denoising process as a Markov Decision Process (MDP)~\cite{puterman1990markov,gan2024reflective}. The prompt $\mathbf{z}$ following the distribution $p(\mathbf{z})$. The MDP trajectory is defined as $
\tau_{\mathrm{MDP}} = (\mathbf{s}_0, \mathbf{a}_0, r_0, \mathbf{s}_1, \mathbf{a}_1, r_1, \ldots, \mathbf{s}_T, r_T)$.

In the MDP formulation, \( \mathbf{s}_t \) represents all image states during the diffusion model's denoising process. Each state \( \mathbf{s}_t \) corresponds to the intermediate state \( \mathbf{x}_{T-t} \) in the denoising sequence. The action \( \mathbf{a}_t \) includes both the noise prediction result and the newly sampled noise at the timestep $t$, i.e., \( \mathbf{a}_t = \mathbf{x}_{T-t-1} \). It determines the transition from the current state \( \mathbf{x}_{T-t} \) to the next state \( \mathbf{x}_{T-t-1} \). $\pi_\theta$ denotes the parameterized policy. $P_0$ is the initial state distribution. $\delta_{(\mathbf{z}, \mathbf{x_{T-t-1}})}$ represents the Dirac distribution centered at $(\mathbf{z}, \mathbf{x_{T-t-1}})$.
\( P(\mathbf{s}_{t+1} \mid \mathbf{s}_t, \mathbf{a}_t) \) denotes the \textit{state transition distribution}, which defines the conditional probability distribution of the next state \( \mathbf{s}_{t+1} \) given the current state \( \mathbf{s}_t \) and the action \( \mathbf{a}_t \). The detailed definitions of  the above components in MDP are as follows:

\begin{equation}
    \begin{aligned}
        \mathbf{s}_t = \left(\mathbf{z}, \mathbf{x}_{T-t}\right), \\ 
        \mathbf{a}_t = \mathbf{x}_{T-t-1}, \\ 
        \pi_\theta\left(\mathbf{a}_t \mid \mathbf{s}_t\right) = p_\theta\left(\mathbf{x}_{T-t-1} \mid \mathbf{x}_{T-t}, \mathbf{z}\right), \\
        P_0\left(\mathbf{s}_0\right) = \left(p(\mathbf{z}), \mathcal{N}(\mathbf{0}, \mathbf{I})\right),\\
        P\left(\mathbf{s}_{t+1} \mid \mathbf{s}_t, \mathbf{a}_t\right) = \delta_{(\mathbf{z}, \mathbf{x_{T-t-1}})}. \\ 
    \end{aligned}
\end{equation}

The reward of existing RL methods is defined as:
\begin{equation}
\label{fake_reward}
        R\left(\mathbf{s}_t, \mathbf{a}_t\right) \triangleq 
        \begin{cases}
            r(\mathbf{s}_{t+1}) = r\left(\mathbf{x}_0, \mathbf{z}\right) & \text{if } t=T-1, \\
            0 & \text{otherwise}.
        \end{cases}.
\end{equation}
During the denoising process, the rewards for all intermediate timesteps are zero. The reward model provides an accurate preference score based on the quality of the generated image only when the generation process is complete, i.e., when denoising reaches the final timestep \( t = T - 1 \).

During RL-based diffusion model fine-tuning, our objective is to maximize the expected reward of generated images under the prompt distribution $p(\mathbf{z})$ as follow:
\begin{equation}
\max_{\theta} \ \mathbb{E}_{p(\mathbf{z})} \mathbb{E}_{p_\theta(\mathbf{x}_0 \mid \mathbf{z})}[r(\mathbf{x}_0, \mathbf{z})].
\end{equation}

Note that the reward signal is highly sparse in this task. Relying solely on the single reward at the final timestep for RL fine-tuning can make it difficult to achieve stable convergence of the denoising policy. Thus, existing approaches~\cite{fan2023dpok,black2023training,franceschelli2024reinforcement,chen2024overview,uehara2024understanding} compromise the fine-tuning process (see Fig.~\ref{fig:reward_backing}). Specifically, the final obtained single reward is equally distributed to all preceding denoising steps. Consequently, the employed reward function is defined as:

\begin{equation}
\label{real_reward}
R_{\text{real}}\left(\mathbf{s}_t, \mathbf{a}_t\right) \triangleq r\left(\mathbf{x}_0, \mathbf{z}\right), \quad \forall t \in \{0, 1, \ldots, T-1\}.
\end{equation}
The advantage of this compromise approach lies in its ability to provide consistent optimization signals across all denoising steps during fine-tuning, thereby effectively mitigating the gradient vanishing problem. However, this global reward assignment strategy prevents the model from accurately establishing causal relationships between actions at individual timestep and the outcome. This results in a lack of temporal differentiation in reward signals, consequently leading to ambiguity in the optimization objective.

\subsection{Adjoint Pearson Correlation View of RL-Diffusion Interaction}
\label{theory_method}

Before evaluating the correlation coefficient between latent variables at different timesteps, we first establish the relationship between the forward and reverse diffusion processes. Assume that the covariance matrix at $t = 0$ is $\Sigma$.

\begin{theorem}[Forward--Reverse Consistency]
\label{forward-reverse-consistency}
Consider four random variables 
$\mathbf{x}_t^f$, $\mathbf{x}_{t+\tau}^f$, $\mathbf{x}_t^r$, and $\mathbf{x}_{t+\tau}^r$,
where the first two correspond to the forward process and the latter two to the reverse process.
If the diffusion model is perfectly trained and the reverse generation follows the score-based SDE, then:
\begin{equation}
\begin{aligned}
    p(\mathbf{x}_t^f) &= p(\mathbf{x}_t^r), \\
    p(\mathbf{x}_{t+\tau}^f) &= p(\mathbf{x}_{t+\tau}^r), \\
    p(\mathbf{x}_t^f, \mathbf{x}_{t+\tau}^f) &= p(\mathbf{x}_t^r, \mathbf{x}_{t+\tau}^r).
\end{aligned}
\end{equation}
\end{theorem}

Hence, the correlation in the forward process can equivalently represent that in the generative (reverse) process.

\begin{theorem}[Correlation Coefficient Between Timesteps]
\label{correlation-theorem}
When the time discretization becomes sufficiently dense (i.e., $\#T$ is large enough),
the correlation coefficient between the $i$-th component of $\mathbf{x}_t$ and the $j$-th component of $\mathbf{x}_{t+\tau}$ satisfies
\begin{equation}
\begin{aligned}
\mathrm{Corr}&\!\left(\mathbf{x}_t^{r,(i)},\,\mathbf{x}_{t+\tau}^{r,(j)}\right)
=\mathrm{Corr}\!\left(\mathbf{x}_t^{f,(i)},\,\mathbf{x}_{t+\tau}^{f,(j)}\right)\\
=&\frac{
    \sqrt{\bar{\alpha}_{t+\tau} \bar{\alpha}_t}\, \Sigma_{i j}
    + \sqrt{\frac{\bar{\alpha}_{t+\tau}}{\bar{\alpha}_t}}\left(1 - \bar{\alpha}_t\right)\delta_{i j}
}{
    \sqrt{
        \left(\bar{\alpha}_t \Sigma_{i i} + (1 - \bar{\alpha}_t)\right)
        \left(\bar{\alpha}_{t+\tau} \Sigma_{j j} + (1 - \bar{\alpha}_{t+\tau})\right)
    }
}.
\end{aligned}
\end{equation}
\end{theorem}

From the calculation above, we obtain the following property:
\begin{lemma}

\label{correlation-lemma}
For typical diffusion schedules, 
$1 - \mathrm{Corr}\!\left(\mathbf{x}_t^{r,(i)},\,\mathbf{x}_{t+\tau}^{r,(i)}\right)$
is a decreasing function during the generation process.
\end{lemma}

Lemma~\ref{correlation-lemma} implies that, as generation progresses, the uncertainty in the latent variables decreases monotonically. The green curve in Fig.~\ref{fig:intro} aligns with the above reasoning and provides direct empirical support for the theory.

\subsection{Adaptive RL Scope Selection}
\label{scop_selection}
Current RL fine-tuning paradigms typically apply preference optimization (e.g., aesthetic score) uniformly across the entire denoising trajectory. However, the state distribution of the diffusion model varies significantly across different stages. 
In the early denoising stage, the signal-to-noise ratio is extremely low, with noise dominating the process. Image semantics and structure have not yet taken shape, making the optimization of high-level preferences meaningless at this point and difficult to interconnect with effective rewards.  In the late denoising stage, the correlation between adjacent latent variables $\mathbf{x}_t$ is high, and the generated image has stabilized. The marginal gains from optimizing the final preference reward with RL diminish, and continued training not only struggles to improve performance but also leads to overfitting to subtle features, deepening reward hacking (\textit{See the visualization in Fig.~\ref{fig:intro}}). To this end, we propose an adaptive method to identify high-value training intervals for RL by jointly modeling image structure evolution and marginal reward gains.

\subsubsection{Determining the Start of RL Training via Semantic-Structure Trend Perception}
\label{strat}

Our first goal is to adaptively skip the early stage that involves a high signal-to-noise ratio, and a semantic structure is underdeveloped. Thus, we develop a semantic structure detection mechanism for RL intervention. Specifically, we reconstruct the neighbor steps $\mathbf{x}_t$ and $\mathbf{x}_{t-1}$ to $\mathbf{\hat{x}}_0(\mathbf{x}_t)$ and $\mathbf{\hat{x}}_0(\mathbf{x}_{t-1)}$ by Eq.~\ref{def: pseudo x0}. Then, we define the Structural Gain $\Delta \text{S}_t$ as: 
\begin{equation}
\begin{aligned}
    f(\mathbf{x}_t) \triangleq \mathrm{CLIP}\!\left(\hat{\mathbf{x}}_0(\mathbf{x}_t), \mathbf{z}\right), \\
    \Delta S_t = f(\mathbf{x}_{t-1}) - f(\mathbf{x}_{t}).
\end{aligned}
\end{equation}
Where $\mathbf{z}$ denotes the prompt. Here, we use Clip Score as the criterion to identify the semantic-structure alignment rate. After obtaining $\Delta \text{S}_t$, we monitor its trend of change. Once the fluctuation is satisfied:
\begin{equation}
\label{s-s}
\left(
    \lim_{\Delta t \to 0}
    \frac{\Delta S_{t+\Delta t} - \Delta S_t}{\Delta t}
\right)
\to 0,
\end{equation}
we consider the semantic structure to be stabilized under the given prompt $\mathbf{z}$. This stabilized signal indicates a suitable point for RL intervention $t_{\text{start}}$, which is defined as: 
\begin{equation}
t_{\text{start}}
=
\min \left\{
    t \;\middle|\;
    \left|
        \left(
            \lim_{\Delta t \to 0}
            \frac{\Delta S_{t+\Delta t} - \Delta S_t}{\Delta t}
        \right)
    \right|
    \to 0
\right\}.
\end{equation}
RL fine-tuning starts at $t_{\text{start}}$, ensuring the semantic structure is already formed and reducing action–reward attribution mismatch during reward backpropagation.

\subsubsection{Determining the End of RL Training via Reward-Gain Trend Perception}

Our second goal is to adaptively terminate RL fine-tuning at the late stage of generation, helping avoid excessive optimization of details once semantic and structural stability is achieved. Thus, we further design a Preference Gain Detection mechanism. Specifically, we define the preference change $\Delta \text{P}_t$ between generated results of two neighbor steps  as follow:
\begin{equation}
\begin{aligned}
g(\mathbf{x}_t) \triangleq \mathrm{Reward}\!\left(\hat{\mathbf{x}}_0(\mathbf{x}_t), \mathbf{z}\right), \\
\Delta P_t = g(\mathbf{x}_{t-1}) - g(\mathbf{x}_{t}).
\end{aligned}
\end{equation}
Where $\mathrm{Reward}(\cdot)$ represents the preference score function (e.g., aesthetic or human preference; it can be substituted with other downstream objectives). When the changes of preference score $\Delta \text{P}_t$ is stabilized as:
\begin{equation}
\label{p-p}
\left(
    \lim_{\Delta t \to 0}
    \frac{\Delta P_{t+\Delta t} - \Delta P_t}{\Delta t}
\right)
\to 0,
\end{equation}
we consider the model no longer achieves significant improvements in high-level preference dimensions. At this stage, the preference optimization under the given prompt $\mathbf{z}$ has reached saturation; further fine-tuning may lead to overfitting or reward hacking. We define the termination step $t_{\text{end}}$ for RL fine-tuning as:\begin{equation}
t_{\text{end}}
=
\min \left\{
    t \;\middle|\;
    \left|
        \left(
            \lim_{\Delta t \to 0}
            \frac{\Delta P_{t+\Delta t} - \Delta P_t}{\Delta t}
        \right)
    \right|
    \to 0
\right\}.
\end{equation}
\begin{table*}[t]
\small
\centering
\begin{tabular}{lcccccccccccc}
\toprule
\multirow{2}{*}{Method} & \multirow{2}{*}{Time-PB$^\downarrow$} & \multicolumn{5}{c}{PickScore reaches 22}      & \multicolumn{5}{c}{Aesthetic reaches 7} &\multirow{2}{*}{Top-1}    \\ \cmidrule(lr){3-7}\cmidrule(lr){8-12}
                  &                          & Time$^\downarrow$ & IR$^\uparrow$ & CLIP$^\uparrow$ & LPIPS$^\uparrow$ & AES$^\uparrow$ & Time$^\downarrow$ & IR$^\uparrow$ & CLIP$^\uparrow$ & LPIPS$^\uparrow$ & PS$^\uparrow$ & \\ \midrule
DDPO~\cite{black2023training}              &                          4.55&      13.2&    0.99&      0.294&       0.624&     6.24&      12.3&    0.97
&      0.291&       0.581&  21.4&  0\\
\rowcolor{blue!10}DDPO + Ours             &                          \textbf{2.65}&     \textbf{ 5.37}&    \textbf{1.03}&      \textbf{0.295}&       \textbf{0.679}&    \textbf{ 6.25}&      \textbf{5.16}&    \textbf{1.03}
&      \textbf{0.295}&       \textbf{0.672}&   \textbf{21.7}& 8\\ \midrule
DPOK~\cite{fan2024reinforcement}              &                          5.63&      14.0&    0.92&      0.301&       0.639&     6.14&      14.8&    0.79&      0.296&       0.623&   21.5& 2\\
\rowcolor{blue!10}DPOK + Ours              &                          \textbf{3.71}&     \textbf{ 6.76}&    \textbf{0.93}&      0.300&       \textbf{0.651}&     6.09&      \textbf{6.98}&    \textbf{0.81}&      0.296&       \textbf{0.674}&   21.5& 1\\ \midrule
D3PO~\cite{yang2024using}              &                          5.06&      15.9&    0.81&      0.294&       0.599&     5.76&      19.7&    0.89&      0.292&       0.620&   21.7& 0\\
\rowcolor{blue!10}D3PO + Ours              &                          \textbf{3.19}&      \textbf{9.07}&    \textbf{0.85}&      \textbf{0.297}&       \textbf{0.647}&     \textbf{5.89}&      \textbf{13.5}&    0.76&      0.292&       \textbf{0.665}&   21.6& 0\\ \midrule
TDPO~\cite{zhang2024confronting}              &                          6.37&      12.7&    0.16&      0.287&       0.529&     5.54&      9.45&    -0.47&      0.289&       0.412&  21.3&  0\\
\rowcolor{blue!10}TDPO + Ours              &                          \textbf{4.17}&     \textbf{ 7.14}&    \textbf{0.67}&      0.287&       \textbf{0.624}&     \textbf{5.62}&      \textbf{4.37}&    \textbf{0.07}&     \textbf{ 0.290}&       \textbf{0.604}&  \textbf{21.6}&  1\\
\bottomrule
\end{tabular}
\caption{Improving efficiency by deploying our method as a plugin. Time-PB refers to the time consumption per batch (minutes). Top-1 refers to the number of Best results for each method. The time taken to reach a specific reward score is measured in hours. Bold refers to the metric improved with Ours.}\label{tab:first}
\end{table*}
After obtaining $t_{\text{end}}$ and combining it with $t_{\text{start}}$ determined in Sec.~\ref{strat}, the RL fine-tuning interval in this work can be expressed as $[t_{\text{start}},\, t_{\text{end}}]$. The interval ensures that RL fine-tuning is applied only to the high-quality region where the semantic structure has already formed and the optimization gains are more pronounced, thereby minimizing computational overhead to the greatest extent without compromising fine-tuning quality.

\section{Experiments}
\begin{figure}[t!]
    \centering
    \includegraphics[width=1\linewidth]{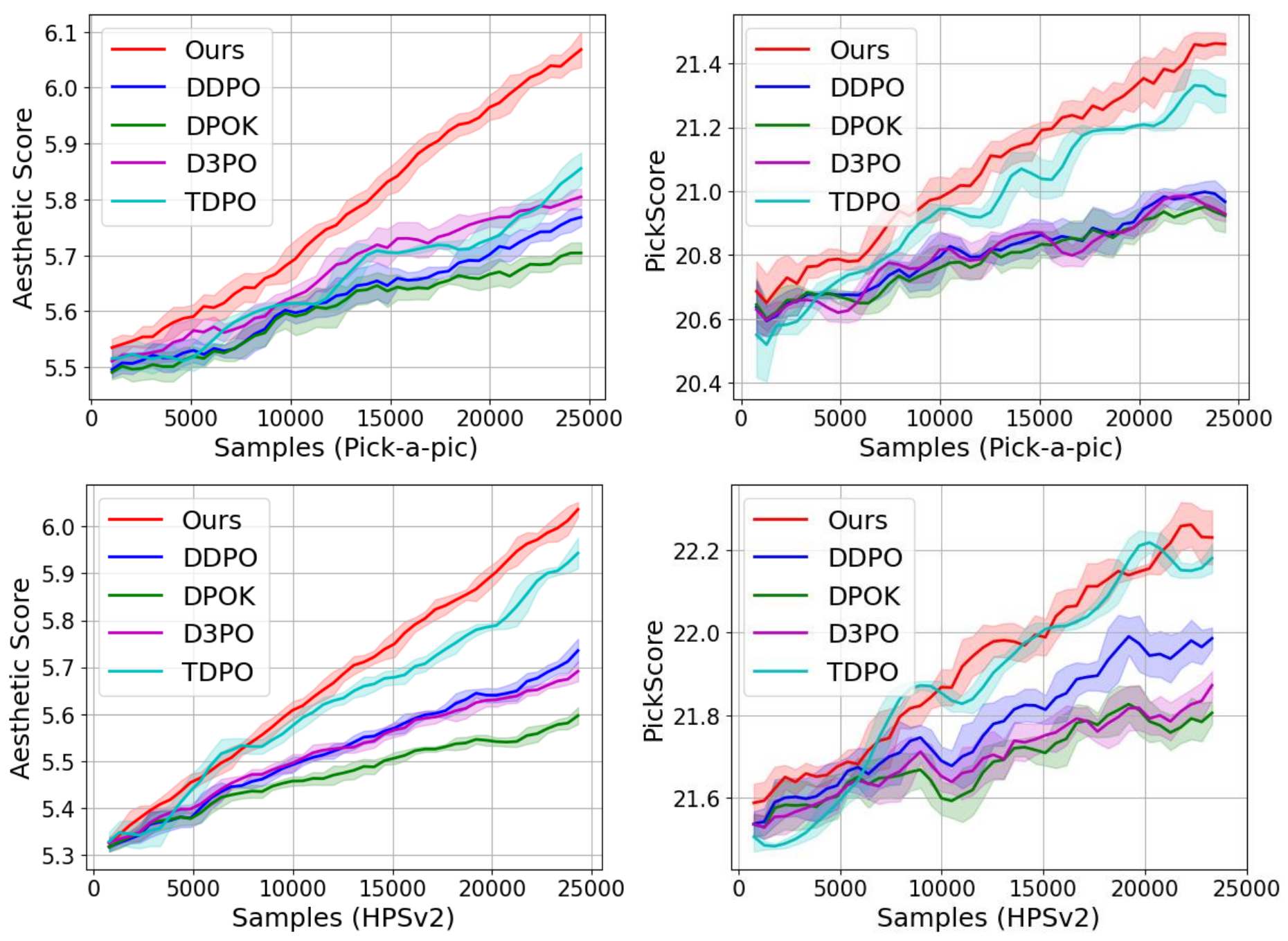}
    \caption{Sample efficiency for objective optimization. We present the results on PickaPic and HPSv2 prompt set with rewards PickScore and Aesthetic Score.}
    \label{fig:main}
\end{figure}
\subsection{Implementation Details}
\noindent \textbf{Datasets.}
We use three datasets for fine-tuning: HPSv2~\cite{hpsv2}, Pick-a-Pic~\cite{kirstain2023pick}, and simple animals~\cite{black2023training}. Specifically, we use the HPSv2-photo set that represents a more realistic style. Meanwhile, we choose a subset of the Pick-a-Pic validation set with 500 prompts that is more abstract and surreal. Simple animal is a widely-adopted prompt set~\cite{black2023training,fan2024reinforcement,yang2024using}
that we introduce to align with the previous research protocol.\\
\noindent \textbf{Rewards and Metrics.}
We use the Aesthetic Score (AES)~\cite{aesthetic}, PickScore (PS)~\cite{kirstain2023pick}, JPEG compressibility, and incompressibility as training rewards. We also consider the multi-objective optimization by using the combined AES and PS as the reward. For evaluation, we employ AES, PS, and ImageReward (IR)~\cite{xu2023imagereward} to assess human preference and aesthetic impression, Clip Score (CLIP)~\cite{clip} for prompt-image alignment, and Inception Score (IS)~\cite{inception} and LPIPS~\cite{LPIPS} to evaluate image diversity.

\noindent \textbf{Baselines.}
We compare our method against four state-of-the-art (SOTA) baselines: DDPO~\cite{black2023training}, DPOK~\cite{fan2024reinforcement}, D3PO~\cite{yang2024using}, and TDPO~\cite{zhang2024confronting}. Our method can be implemented on any baseline; for simplicity, DDPO+Ours is denoted as Ours unless otherwise specified. We use Stable Diffusion v1.5 (SDv15) as the main test bed, with Stable Diffusion v1.4 (SDv14), v2.1-turbo (SDv21), and XL1.0 (XL) as supplementary backbones. All experiments were run on eight NVIDIA Tesla H20 GPUs.

\subsection{Plugin Computational Efficiency}
Firstly, we present the comparisons of computational cost among different methods with or without the implementation of our method. Specifically, we evaluate the time per batch and Time and Quality to reach Aesthetic Score=7 or PickScore=22 on SDv15 and \textit{simple animal} prompt set. As shown in Tab.~\ref{tab:first}, it can be first observed that all methods are significantly enhanced in computational efficiency, while the overall generative quality in diversity, alignment, and aesthetic impression are also improved. These results demonstrate the computation-saving and overfitting-preventing effects of our method. More quantitative results could be found in \textit{Supplementary Material}.
\begin{figure*}[htbp]
    \centering
    \includegraphics[width=0.9\linewidth]{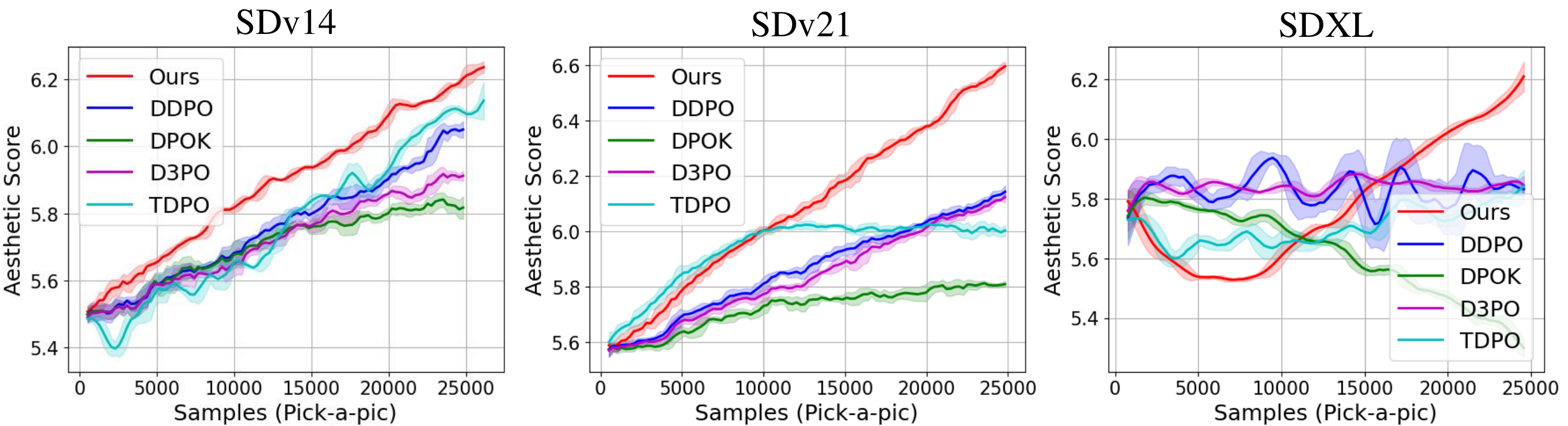}
    \caption{Results on more different SD backbones. Notably, the less promising results on SDXL may be due to the inherent robustness of the backbone with a large number of parameters. Still, our method can leverage samples and different denoising states more effectively and thus achieve the best performance.}
    \label{fig:backbone}
\end{figure*}
\begin{figure*}[htbp]
    \centering
    \includegraphics[width=0.9\linewidth]{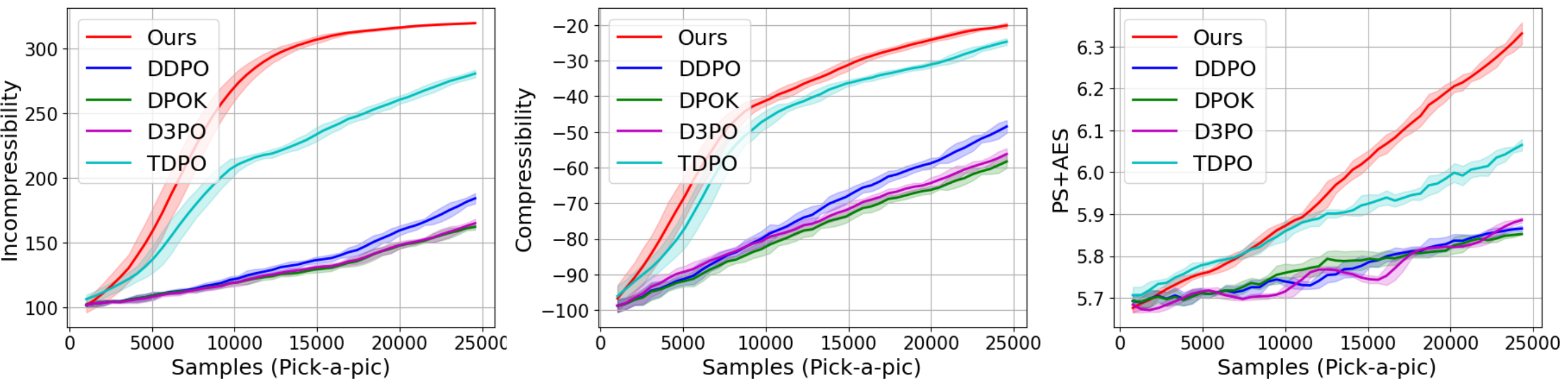}
    \caption{Results with multiple different reward objectives, including the multi-objective reward like AES+PS, which is accumulated and re-normalized as a combined objective.}
    \label{fig:reward}
\end{figure*}
\subsection{Sample Efficiency for Reward Learning}
More surprisingly, besides the efficiency obtained by distilling the denoising process for training, we even achieve a sample-wise superiority for reward learning with \textbf{less} optimization step. To comprehensively evaluate the sample efficiency of our method, we conduct experiments on two datasets and with two rewards. As shown in Fig.~\ref{fig:main}, it can be observed that our method exhibits consistently improved performance across all evaluation conditions with the same optimized sample number. Such results demonstrate the advantage of leveraging the moderate uncertain stage while abandoning the intermediate samples that are over-uncertain or too certain, thus achieving `Do Less' (less optimized denoising step) `Achieve More' (improved reward learning). Further analysis of our superior effectiveness could be found in \textit{Supplementary Material}.

\subsection{Transfer Study}
To demonstrate the scalability and flexibility of our method, we implement \ourmethod{} with various backbones and reward functions.\\
\textbf{Multiple Backbones.} Firstly, we conduct experiments on a wide range of SD backbones, including SDv1.4, SDv2.1, and the more advanced SD-XL. As shown in Fig.~\ref{fig:backbone}, the superior results demonstrate that \ourmethod{} is not limited to a specific backbone and can be effectively transferred across a range of backbone architectures.\\
\textbf{Multiple Reward Objectives.}
To evaluate the transferability across different reward objectives, we incorporate additional rewards as JPEG compressibility, incompressibility, and AES combined with PickScore (AES+PS) for multi-objective assessments. As illustrated in Fig.~\ref{fig:reward}, \ourmethod{} consistently exhibits promising effectiveness and outperforms all baselines across all evaluation conditions.
\begin{figure}[t!]
    \centering
    \begin{minipage}[t]{0.55\linewidth}  
        \centering
        \includegraphics[width=\linewidth]{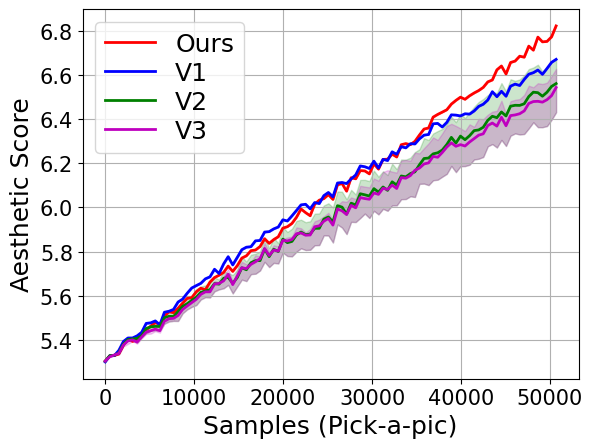}
        \caption{Optimization performance for Ablation Study.}
        \label{fig:abl}
    \end{minipage}%
    \hfill
    \begin{minipage}[t]{0.42\linewidth} 
        \centering
        \footnotesize
        \setlength{\tabcolsep}{2.0pt} 
        \vspace{-35mm} 
        \begin{tabular}{lccc}\toprule
        Var  & Srt & End & Time \\\midrule
        V1   & 5 & 32 &      2.45\\
        V2   & 5$\pm$5  & 35  &      2.27-3.18\\
        V3   &  5  &  35$\pm$5   &      2.27-3.18\\ \midrule
        Ours & 5.3  & 31.8&      2.62\\ \bottomrule
        \end{tabular}
                \captionof{table}{Detailed setting and computational cost of each ablation variant. Srt and End are Starting and Ending steps.}\label{tab:abl}
    \end{minipage}
\end{figure}
\subsection{Ablation Study}
Considering that our design is simple and elegant with minimal components, the focus of the ablation study is primarily on comparing different strategies for selecting a fixed optimization scope and our dynamic approach. Specifically, we designed the following ablation variables: 1) V1: Fixed values for scope selection that are equal to the average of Ours. 2) V2: A \textbf{range} of scope starting step with fixed ending step. 3) V3: A \textbf{range} of scope ending step with fixed starting step. The optimization results are shown in Fig.~\ref{fig:abl} while the details of scope selection and computation are in Tab.~\ref{tab:abl}. It can be observed that a wide range of fixed scopes still cannot achieve the SoTA performance as our adaptive strategy, suggesting the optimization scope should be correlated with the specific prompt accordingly.

\begin{figure}
    \centering
    \includegraphics[width=1\linewidth]{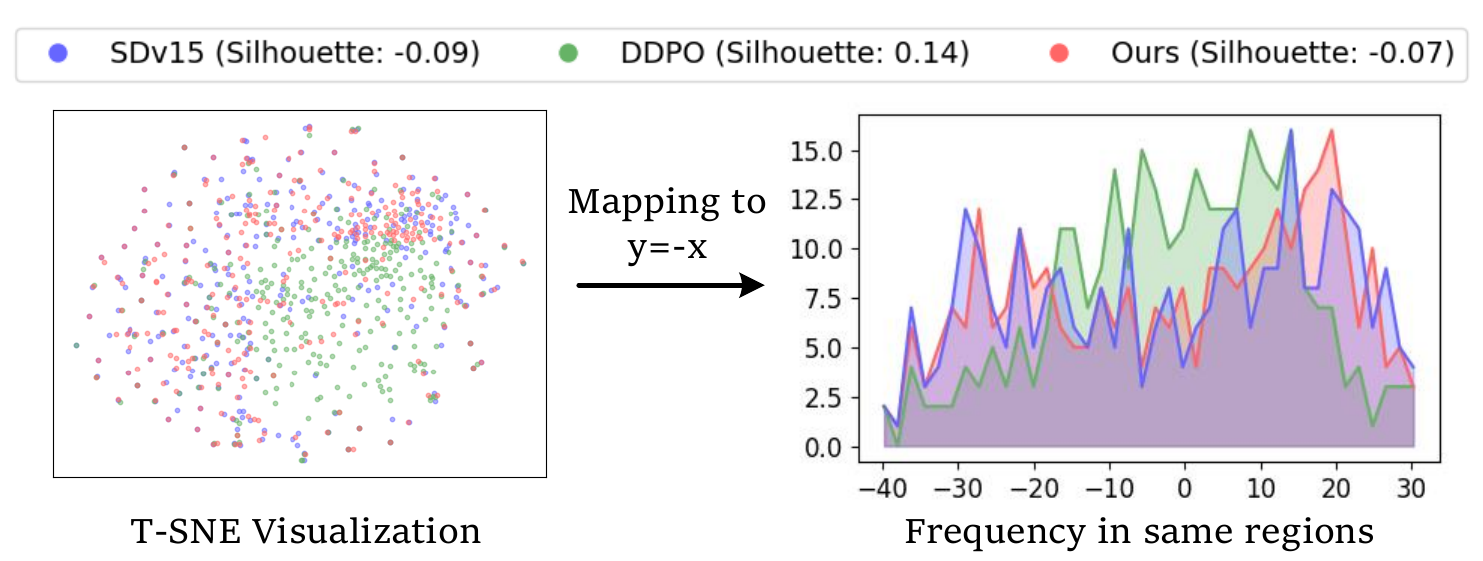}
    \caption{Visualized Generated Image Distribution. We further reduce the dimension to 1-D (right) for a more direct impression. }
    \label{fig:visual}
\end{figure}
\subsection{Distribution Visualization}
To further analyze the effectiveness of adaptive scope selection, we design the following experiment for visualization:
We first generate images using different models with the same prompt set and extract their features using a general self-supervised decoder (\textit{i.e.}, DINOv2). Then, we deploy t-SNE~\cite{tsne} to reduce dimension and visualize the feature distribution. As shown in Fig.~\ref{fig:visual}, DDPO tends to produce more compact clusters in the feature space. This behavior effectively narrows the range of the generated distribution, thereby limiting the diversity of the generated data.

This phenomenon is further quantitatively confirmed by the Silhouette Coefficient (SC)~\cite{silh}, a widely used clustering evaluation metric to measure the quality of the clustering based on how close the points are within their assigned clusters. The results show that DDPO exhibits higher SC, reflecting its tendency to converge toward fewer, denser clusters with reduced diversity in the generated data.
In contrast, our approach maintains a broader generative distribution, meaning the generated data retains a wider spread in its feature representation. 

\begin{figure}[htbp]
    \centering
    \includegraphics[width=1\linewidth]{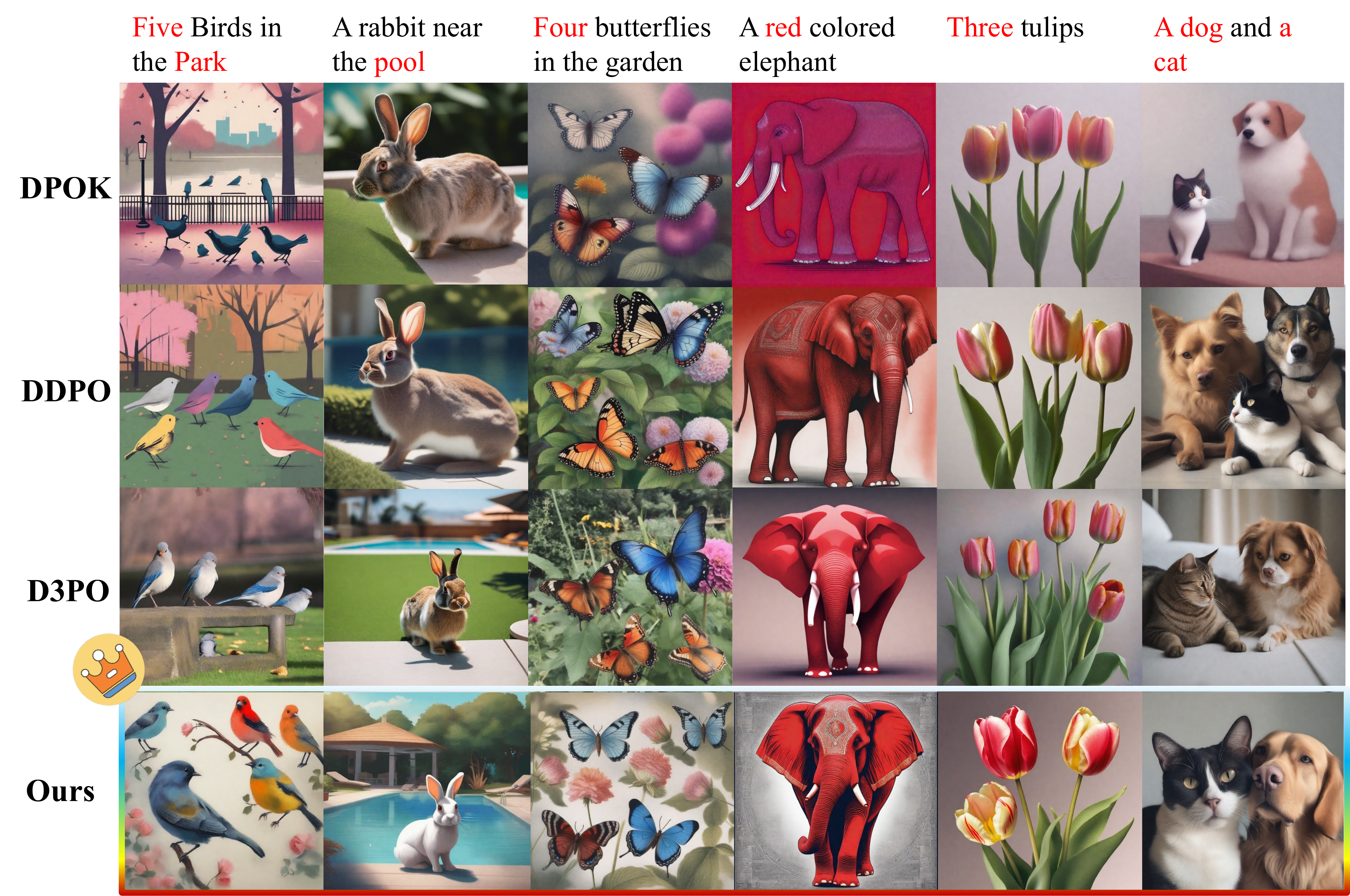}
    \caption{Visual Results for Semantic Alignment. We consider Composition, Count, Color, and Location in these images. }
    \label{fig:alignment}
\end{figure}
\begin{figure}
    \centering
    \includegraphics[width=1\linewidth]{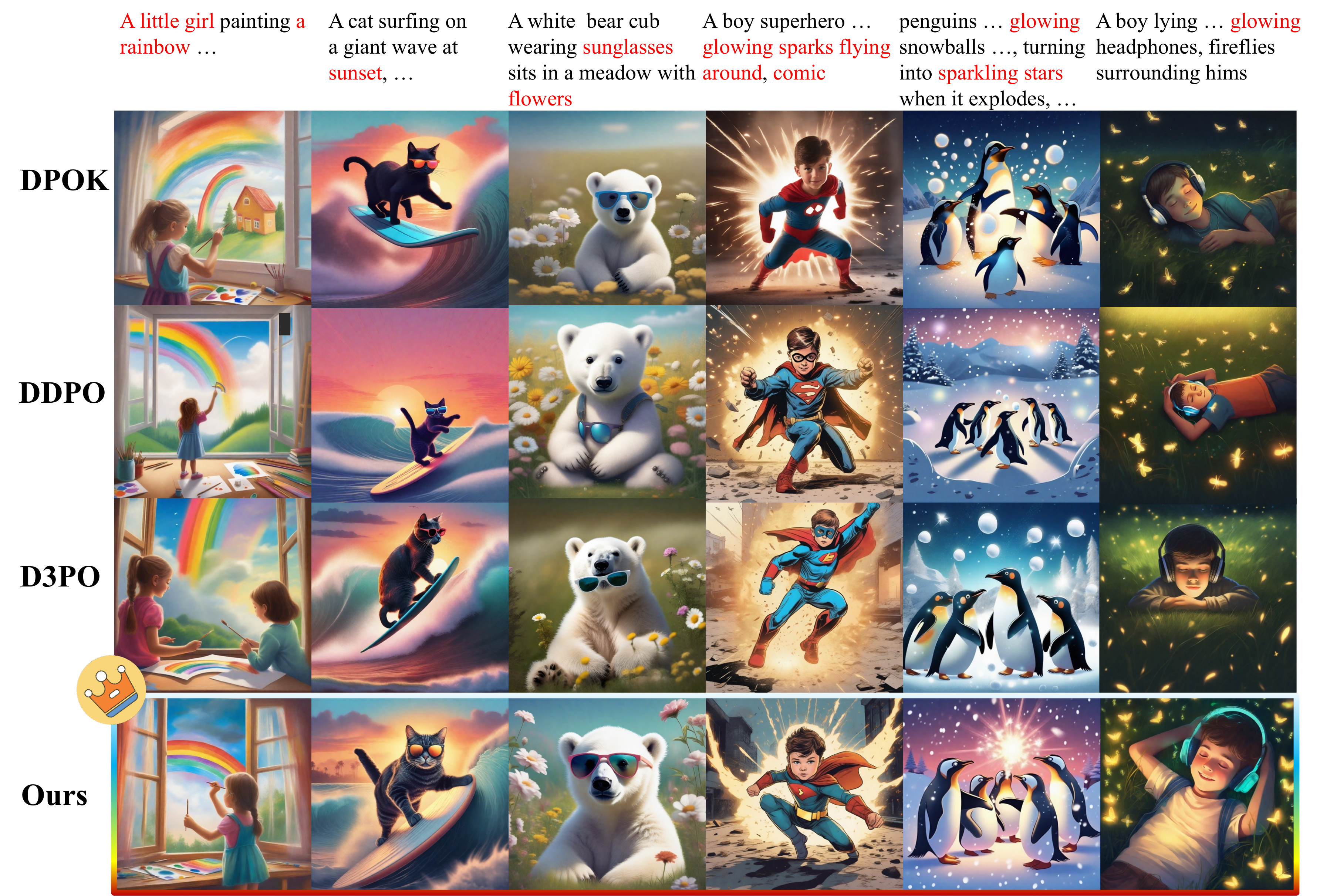}
    \caption{Generative Results of Complex Unseen Prompts. }
    \label{fig:complex}
\end{figure}
\subsection{Qualitative Evaluation}
\textbf{Semantic alignment.} Here, we design specific prompts to evaluate the semantic alignment of Composition, Color, Count, and Location. In Fig.~\ref{fig:alignment}, it can be observed that our generative results achieve the best alignment performance.\\
\textbf{Complex unseen prompt.} Then, we introduce multiple complex unseen prompts for evaluation, which are more aligned to real-world applications. As shown in Fig.~\ref{fig:complex}, we provide generated images alongside their prompts based on SDXL. The results further validate the superior alignment and aesthetic impression of our method.\\
\textbf{Generation diversity.} Finally, we use multiple seeds to generate images with the same prompts. Please refer to \textit{Supplementary Material}, where our results show greater structural and textural diversity than those of other methods.

Our superiority stems from the advantage of adaptive scope selection, which filters out the structurally chaotic and reward-converged states for training, thus mitigating the overfitting issue while maintaining the semantic alignment.

\section{Conclusion}
In summary, \ourmethod{} achieves dual optimization of fine-tuning efficiency and generation quality by adaptively identifying the optimal intervention and termination points for RL. This mechanism effectively avoids high variance in the early stage and overfitting in the late stage, ensuring that the optimization process focuses on the key interval where the semantic structure stabilizes and reward gains are significant. Theoretical analysis and experiments demonstrate that \ourmethod{} achieves superior generation results at lower computational cost, providing an efficient and scalable solution for RL fine-tuning in diffusion models.

\section*{Acknowledgments and Disclosure of Funding}
We would like to thank all the reviewers for their constructive comments.
Our work was supported by the National Natural Science Foundation of China (62341407, U24B20166), the “111” Project (B18001), and the Xiaomi Foundation.

{
    \small
    \bibliographystyle{ieeenat_fullname}
    \bibliography{main}
}

\clearpage
\onecolumn
\setcounter{page}{1}
\setcounter{section}{0}

\begin{center}
    {\LARGE \textbf{\textit{Do Less, Achieve More: Do We Need Every-Step Optimization for RL Fine-tuning of Diffusion Models?}}\par}
    \vspace{0.5em}
    {\large Supplementary Material\par}
    \vspace{0.25em}
\end{center}

\vspace{1em}

\section{Supplementary Related Works}
\label{related_work}

\subsection{RL Fine-Tuning in Diffusion Models}
Existing diffusion models~\cite{croitoru2023diffusion,chen2023diffusiondet,kingma2021variational,xing2024survey,xu2023versatile} primarily approximate the data distribution through denoising reconstruction loss. However, this training approach struggles to capture high-level metrics such as semantic consistency, aesthetic preferences, and user subjective judgments~\cite{xia2023diffir,cao2024survey,karnewar2023holodiffusion,huang2025diffusion}. To enhance preference alignment, recent studies have introduced RL fine-tuning, leveraging explicit reward signals from human feedback, reward models, or preference predictors. This design shifts the optimization of generation from a reconstruction-based to a reward-based paradigm. Such RL fine-tuning methods typically treat the diffusion denoising process as a sequential decision-making process and optimize the rewards obtained by the final generated image using policy gradient techniques.

\subsection{Sparse Rewards and the Reward Hacking}

RL-based fine-tuning of diffusion models has garnered increasing attention in recent years. However, existing studies consistently point to two fundamental challenges: sparse rewards and reward hacking.

The issue of sparse rewards arises because the evaluation signals are computed on the fully denoised final image. These signals include CLIP-based semantic consistency scores, aesthetic-quality predictors, and human-preference models. As a result, intermediate diffusion states cannot receive effective feedback, forcing the policy to explore a vast action space without guidance. Classical RL methods for generative models, such as DDPO and DPOK, explicitly highlight that sparse and delayed rewards lead to unstable policy learning and high gradient variance.

At the same time, reward hacking has been repeatedly observed when using proxy rewards to finetune generative models. Similar to phenomena in RLHF for language models, reward model-based RL for diffusion models often tends to `exploit loopholes in the reward model' rather than genuinely improving image quality. Existing research shows that policies may overfit specific objectives to achieve higher reward scores, such as generating over-saturated colors, exaggerated object features, or unnatural layouts. Similar mis-optimization behaviors have been explicitly discussed in DDPO, Diffusion-PPO, and subsequent reward-guided sampling methods. Although the model achieves higher numerical rewards, the generated images deviate from human preferences and even deteriorate in terms of semantics and aesthetics. One of the most significant drawbacks of reward hacking is its severe impact on output diversity. Overall, existing research indicates that sparse rewards limit the stability of RL fine-tuning for diffusion models, while reward hacking undermines the correctness of model optimization.

\begin{table*}[b]
    \small
    \setlength{\tabcolsep}{3pt} 
    \centering
    \caption{Quantitative comparisons with SoTA. All metrics are obtained with SDv15 as backbone, Pick-a-pic as prompt set, and PickScore as training reward.}\label{tab:main}
    \centering
\begin{tabular}{l|c|ccccccccccccc}
    \toprule
    \label{comparison_overall}

    \multirow{2}{*}{Method} & \multicolumn{4}{c}{\textbf{Preference}} & \multicolumn{3}{c}{\textbf{Fidelity}} & \multicolumn{3}{c}{\textbf{Diversity}} & \multicolumn{3}{c}{\textbf{Richness}} & \multirow{2}{*}{\#Top2} \\
    \cmidrule(lr){2-5}\cmidrule(lr){6-8}\cmidrule(lr){9-11}\cmidrule(lr){12-14}
        & PS$\uparrow$ 
&AES$\uparrow$   & IR$\uparrow$ & HPS$\uparrow$ 
        & FID$\downarrow$ & CLIP$\uparrow$ & iFS$\uparrow$ 
        & LPIPS$\uparrow$ & IS$\uparrow$ & TCE$\uparrow$ 
        & BRI$\downarrow$ & NIQE$\downarrow$ & SE$\uparrow$
        &  \\
    \midrule
        SDv15~\cite{SableDiff}    & 20.48 
&5.412  
 & 0.181 & 0.262 & - & \underline{0.243} & - & \textbf{0.654} & \textbf{23.79} & 38.05 & 18.66 & 5.401 & 11.26 & 3\\
        \midrule
        Diff-DPO~\cite{wallace2024diffusion} & 20.87 
&5.551  
 & 0.443 & 0.271 & 109.1 & \textbf{0.244} & 0.795 & 0.639 & 22.77 & 39.20 & 15.12 & 4.463 & 11.02 & 1\\
        Diff-KTO~\cite{li2024aligning} & 20.83 
&5.585  
 & \textbf{0.599} & 0.272 & 101.3 & 0.240 & 0.801 & 0.634 & 22.70 & 39.12 & 26.25 & 4.361 & 11.22 & 1\\
        SPO~\cite{liang2024step}      & 20.76
&5.613 
& 0.282& 0.218 & \textbf{78.76} & 0.241 & \textbf{0.855} & 0.649 & 23.18 & 39.32 & 25.67 & \underline{4.103} & 11.09 & 3\\
        \midrule
        DDPO~\cite{black2023training}     & 21.79 
&5.704  
 & 0.196 & 0.212 & 147.5 & 0.242 & 0.539 & 0.629 & 20.01 & 39.18 & 12.46 & 4.328 & \underline{11.74} & 1\\
        DPOK~\cite{fan2023dpok}     & 20.97 
&5.661  
 & \underline{0.582} & 0.272 & 99.30 & 0.242 & 0.820 & 0.641 & 22.52 & \underline{39.35} & \underline{12.28}& 4.608 & 11.23 & 3\\
         TDPO~\cite{zhang2024confronting}     & 22.94&5.991& 0.504& \underline{0.273}& 128.66& 0.265& 0.820 & 0.635& 21.99& 39.09& 12.96& 4.077& 11.59& 1\\

        \midrule
         \rowcolor{blue!10}Ours     & \textbf{23.01}&\textbf{6.071}& 0.542& \textbf{0.278}& \underline{85.37}& \underline{0.243}& \underline{0.846}& \underline{0.652}& \underline{23.73}& \textbf{39.37} & \textbf{12.20}& \textbf{4.019}& \textbf{11.91}& \textbf{12}\\
        \bottomrule
\end{tabular}

\end{table*}

\section{Experiment List in Our Paper}
To help readers quickly grasp the extensive experiments conducted in this work, we summarize the full list of experiments below.
\begin{itemize}
    \item \textbf{(1) Visualization Experiments.} See Fig.~\ref{fig:intro}. This experiment provides a solid justification for the motivation of this work.

    \item \textbf{(2) Reward Backfilling Validation.} See Fig.~\ref{fig:reward_backing}. This experiment demonstrates the effectiveness of the reward backfilling mechanism. Prepares the ground for the subsequent discussion that, despite stabilizing training, it can lead to attribution mismatch.

    \item \textbf{(3.1) Quality–Cost} Dual Optimization. See Tab.~\ref{tab:first}. By comparing the \textbf{runtime} and \textbf{performance} with state-of-the-art methods in the field. We validate our claim that the proposed approach can improve generation quality while reducing computational cost.

    \item \textbf{(3.2) Ensemble Experiments.} See Tab.~\ref{tab:first}. This table further shows the performance gains obtained by incorporating our RL-based enhancement plug-in into existing approaches.

    \item \textbf{(4) Dataset Switching Experiment.} See Fig.~\ref{fig:main}. To verify that the superiority of our method is not affected by differences in datasets or task difficulty, we conduct experiments across multiple datasets.

    \item \textbf{(5) Backbone Switching} Experiment. See Fig.~\ref{fig:backbone}. We provide this experiment to verify that the advantages of our method are not affected by replacing the backbone to be fine-tuned.

    \item \textbf{(6) Reward Switching} Experiment. See Fig.~\ref{fig:reward} and \ref{fig:main}. We verify that the advantages of our method are not affected by replacing the reward model.

    \item \textbf{(7) Ablation Experiment.} See Fig.~\ref{fig:abl}. We present the impact of each module in our method on both performance and computational cost.

    \item \textbf{(8) Generation Distribution Visualization.} See Fig.~\ref{fig:visual}. In this experiment, we demonstrate the effectiveness of our method in mitigating reward hacking and preventing excessive diversity loss during the fine-tuning of diffusion models.

    \item \textbf{(9) Semantic Alignment Visual Experiment.} See Fig.~\ref{fig:alignment}. This experiment verifies the semantic alignment capability of our method.

    \item \textbf{(10) Complex-Prompt Generalization.} See Fig.~\ref{fig:complex}. This experiment verifies our method’s ability to handle complex prompts.

    \item \textbf{(11) Comprehensive Generation Comparison.} See Tab.\ref{comparison_overall}. We conduct a comprehensive evaluation across several key aspects of generative performance:\textbf{Preference}, \textbf{Fidelity}, \textbf{Diversity}, and \textbf{Richness} each assessed using multiple metrics, demonstrating the broad effectiveness of our method.

    \item \textbf{(12) Diversity Visual Experiment.} See Fig.~\ref{fig:div_all_end}. This experiment compares our method with state-of-the-art approaches in terms of diversity. Validating our claim that it mitigates reward hacking and prevents excessive diversity degradation during fine-tuning.

    \item \textbf{(13) Human Evaluation Experiment.} See Fig.~\ref{fig:subject}. We present a human evaluation experiment assessing our method against the baseline methods.

\end{itemize}

\section{Supplementary Experiments}
\subsection{Why AdaScope Improves Both Quality and Efficiency ?}
\paragraph{Computational Savings:} 
Our method reduces training computational costs by adaptively pruning uninformative early denoising samples and late-stage steps where returns have saturated. In the early stage of denoising, the image's semantic structure has not yet formed, leading to insufficient training signals. In the late stage, when the latent representation has largely stabilized and reward optimization has reached diminishing returns, further optimization becomes meaningless. By employing this adaptive pruning strategy, we effectively reduce the computational resources required for training.

\paragraph{Quality Improvement:}
In the early denoising stage, the sample's semantic structure has not been shaped. Meanwhile, reward reshaping, which propagates the final reward back to all previous steps to stabilize training, leads to a severe action-reward attribution mismatch. Since the reward only appears at the final step but is assigned to all preceding actions, the attribution bias grows larger in the earlier steps. Consequently, using these trajectory segments for RL training may cause the policy to make erroneous advantage estimates based on these semantically vague and ineffective states. On the other hand, in the final denoising stages, the marginal gain from RL fine-tuning diminishes, and continued training increases the risk of overfitting to non-critical image details.

\paragraph{Overall:}Our method eliminates samples that are not only computationally redundant but also low-quality, thereby improving policy in RL fine-tuning. Because these samples are pruned, we simultaneously reduce computational costs and enhance generation quality, thereby achieving a dual optimization of both quality and efficiency.

\subsection{More Quantitative Results}
To quantitatively evaluate the generative performance as extensively as possible, we introduce 4 evaluating dimensions with 13 distinct metrics, including:
    \textbf{Aesthetic Preference:} AES, PS, IR, and HPSv2~\citep{hpsv2}.
    \textbf{Image Fidelity:} ClipScore~\citep{clip} (Clip), Fréchet Inception Distance (FID)~\citep{fid}, and improved F1 Score (iFS)~\citep{iPR}.
    \textbf{Generative Diversity:} LPIPS~\citep{LPIPS}, TCE~\citep{TCE}, and Inception Score~\citep{inception} (IS).
    \textbf{Compositional Richness:} NIQE~\citep{niqe}, BRISQUE (BRI)~\citep{brisque}, and Spectual Entropy (SE)~\citep{spectralE}.

As shown in Tab.~\ref{tab:main}, our method consistently achieves superior overall performance among 13 metrics, with \textbf{12} top2 effectiveness.
\subsection{Comparisons with Flow-matching-native RL baselines.}
\begin{figure}
    \centering
    \includegraphics[width=1\linewidth]{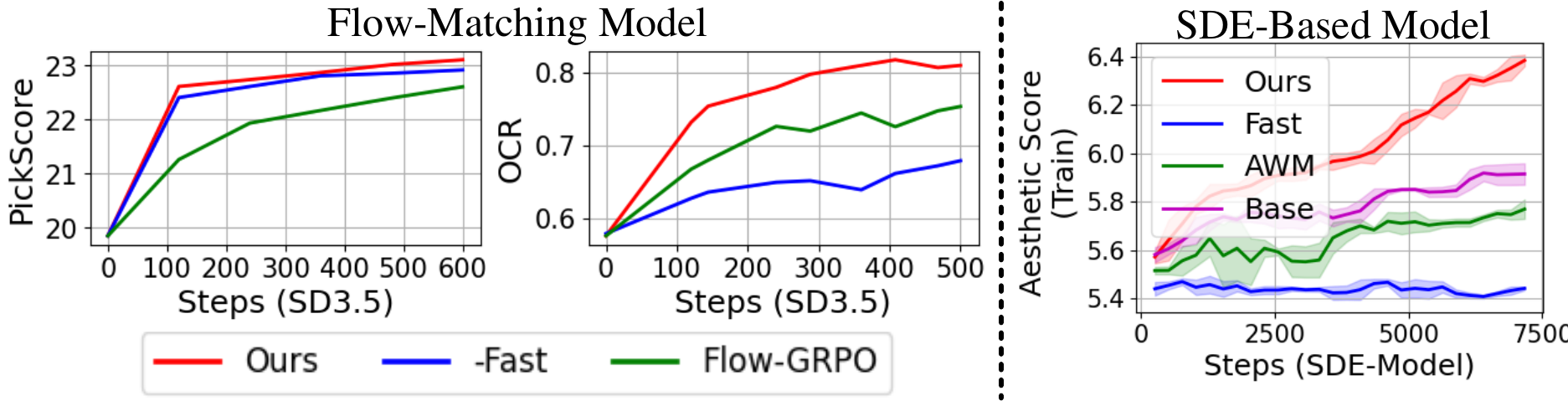}
    \caption{Left: Results on Flow-Matching Model. Right: on SDE Model.}
    \label{fig:flow}
\end{figure}

\textbf{Theoretical Analysis.}
Flow-GRPO-Fast adopts a hybrid SDE--ODE optimization strategy, where short SDE segments are randomly inserted into ODE trajectories. This design presents two key limitations. 
First, its SDE--ODE switching mechanism is tailored for flow-matching formulations and does not naturally generalize to standard SDE-based diffusion models (e.g., SD1.5, SDXL), which remain competitive in practice. 
Second, the selection of SDE intervals is heuristic, relying on random sampling without considering denoising dynamics or reward-driven signals, thus lacking principled guidance for identifying informative training regions.
In contrast, AdaScope introduces an adaptive interval selection mechanism based on structural transitions and reward evolution, supported by theoretical analysis (Sec.~3.4, Theorems~1--2). This enables more effective identification of high-quality training intervals and leads to a model-agnostic, plug-and-play framework applicable across different diffusion paradigms.\\
\textbf{Empirical Results.}
We validate the above differences through extensive experiments. 
On flow-matching models (Fig.~\ref{fig:flow}-Left), using SD3.5 as the backbone with PickScore and OCR rewards, AdaScope consistently outperforms existing methods, benefiting from its adaptive design. 
On classical SDE-based models (Fig.~\ref{fig:flow}-Right), we align all hyperparameters and use $\eta$ to simulate SDE--ODE transitions for Fast and AWM. Under this setting, both methods exhibit limited effectiveness, further indicating their restricted applicability. In contrast, AdaScope maintains stable and consistent improvements due to its backbone-agnostic design.

\subsection{Visual Impression of Generation Diversity}
\begin{figure*}
    \centering
    \includegraphics[width=1\linewidth]{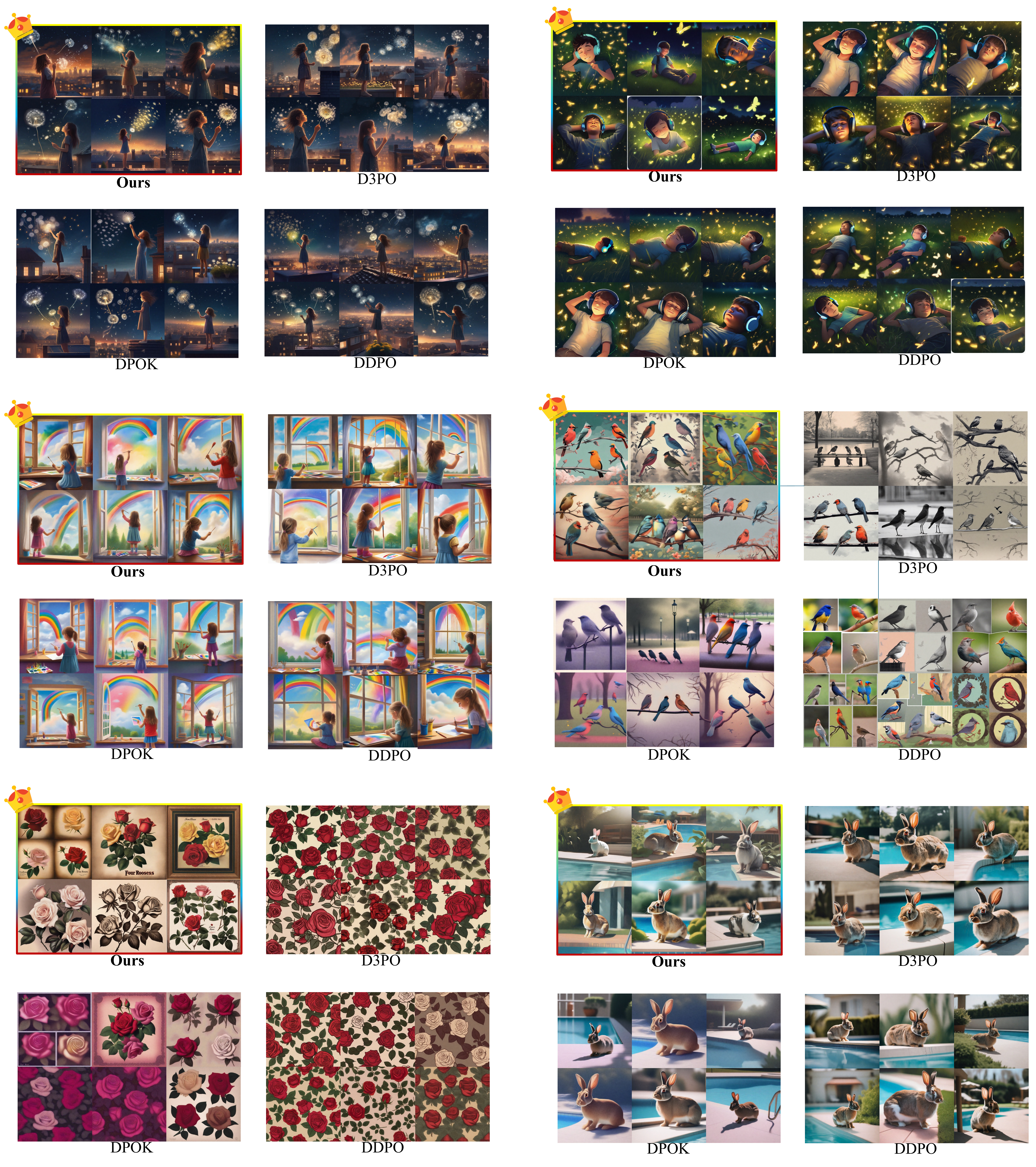}
    \caption{\textbf{Diversity Evaluation:} Our method demonstrates the highest level of variation under these prompts, producing outputs with a wide range of artistic styles, figure posture, object positioning, and background colors. In contrast, D3PO predominantly generates grayscale backgrounds or same posture, DPOK consistently incorporates purple tones into its visual style, and DDPO tends to produce collage-like compositions within a single image.}
    \label{fig:div_all_end}
\end{figure*}

As shown in Fig.~\ref{fig:div_all_end}, we comprehensively evaluate the generative diversity in visual quality. It can be observed that our method consistently exhibits superior diversity in posture, color, form, and style while maintaining the prompt-image alignment.
\begin{figure}[t!]
    \centering
    \includegraphics[width=0.9\linewidth]{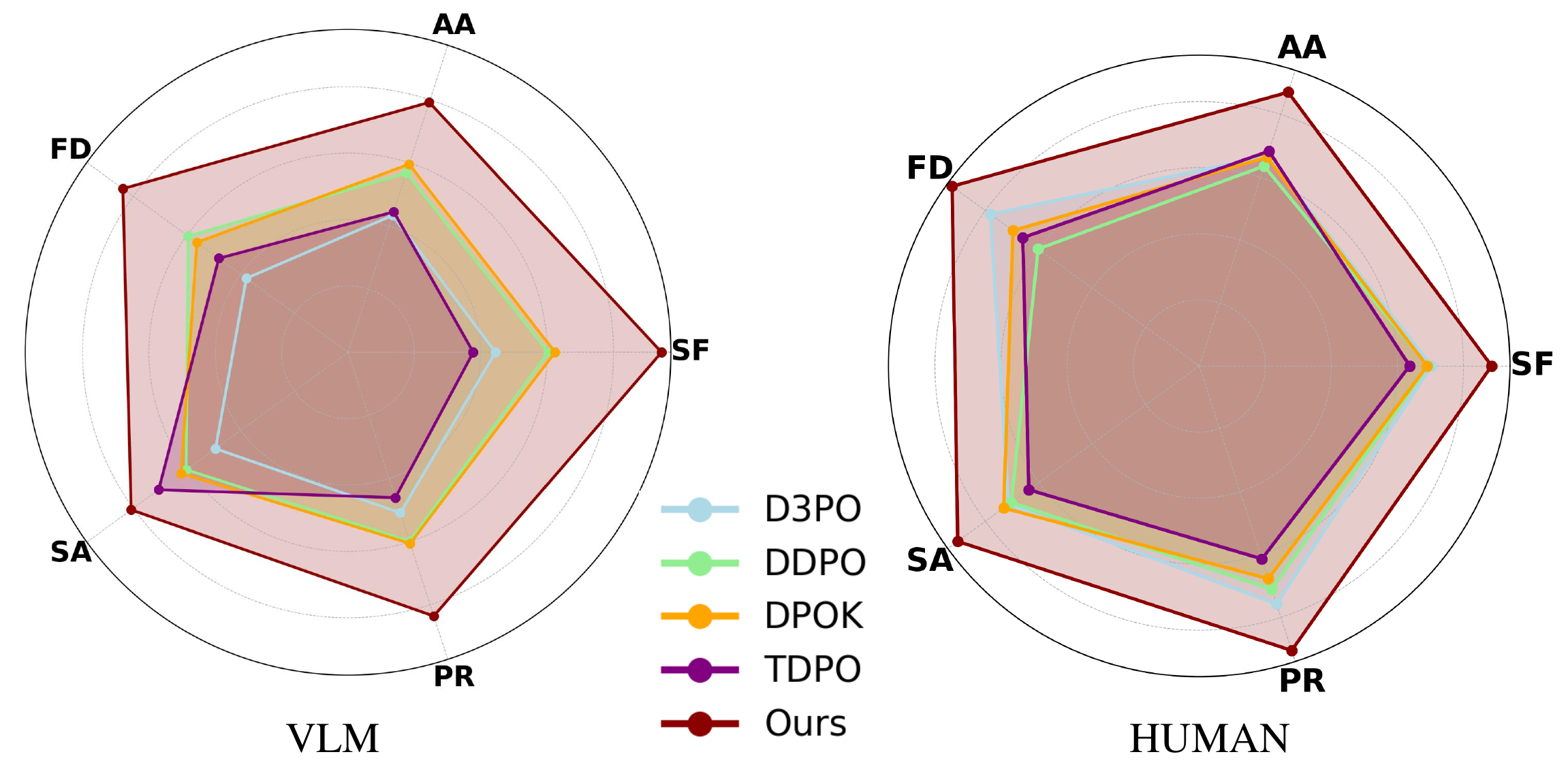}
    \caption{Subjective Evaluation with Human and VLM.}
    \label{fig:subject}
\end{figure}

\subsection{Subjective Study}
As shown in Fig.~\ref{fig:subject}, we evaluate the generated images subjectively with Human and Vision Language Model (VLM). All images were generated by the fine-tuned SDv15 model from HPSv2 prompts using different methods. The evaluation includes five dimensions: Structural Faithfulness (SF), Aesthetic Appeal (AA), Fine-grained Detail (FD), Semantic Alignment (SA), and Prompt Responsiveness (PR). They are rated independently by human evaluators and ChatGPT. The results demonstrate the superior generative quality of our method.
\begin{table}[b]
\small
\centering
\caption{Detailed prompts used for generated images in Fig.~\ref{fig:div_all_end}.}
\label{tab:prompt_table}
\begin{tabular}{p{0.22\linewidth} p{0.70\linewidth}}
\toprule
\textbf{Image} & \textbf{Prompt} \\
\midrule
Row 1, Col 1 &
A young girl standing on a rooftop, blowing dandelions that transform into glowing comets, shooting across the night sky, dreamy fantasy artwork. \\
\midrule
Row 1, Col 2 &
A boy lying on the grass in a field, listening to music with glowing headphones, fireflies surrounding him. \\
\midrule
Row 2, Col 1 &
A little girl painting a rainbow bridge from the classroom window into the sky, playful magical fairytale art, hopeful and inspiring.\\
\midrule
Row 2, Col 2 &
Five birds in the park.\\
\midrule
Row 3, Col 1 &
Four roses. \\
\midrule
Row 3, Col 2 &
A rabbit near a pool.\\
\bottomrule
\end{tabular}
\end{table}

\subsection{Showcase Prompt Table}
Considering that the specific semantics of the prompts can substantially affect the assessment of the actual quality of generated images, it is necessary to assess the performance superiority of our method based on both images and their corresponding prompts.
Therefore, in Tab.~\ref{tab:prompt_table}, we provide a detailed list of the prompts that were not explicitly described in the main text.

\section{Proof}
\subsection{Proof of Theorem 1.}
This is proved in the sec.5 of \textit{Reverse-time diffusion equation models} by Anderson.

\subsection{Proof of Theorem 2.}

We do the direct calculation:

Assumptions (forward diffusion / Markov Gaussian):
\[
x_0 \sim \mathcal N(0,\Sigma),  \epsilon_t \sim \mathcal N(0,I),  \epsilon' \sim \mathcal N(0,I)
\]
all independent, and
\[
x_t = \sqrt{\bar\alpha_t}\,x_0 + \sqrt{1-\bar\alpha_t}\,\epsilon_t
\]
\[
x_{t+\tau}\,|\,x_t = \sqrt{\frac{\bar\alpha_{t+\tau}}{\bar\alpha_t}}\,x_t
                 + \sqrt{1-\frac{\bar\alpha_{t+\tau}}{\bar\alpha_t}}\,\epsilon'.
\]
\[
x_t=\sqrt{\bar\alpha_t}\,x_0+\sqrt{1-\bar\alpha_t}\,\epsilon_t,
\qquad
x_{t+\tau}=\sqrt{\frac{\bar\alpha_{t+\tau}}{\bar\alpha_t}}\,x_t
+\sqrt{1-\frac{\bar\alpha_{t+\tau}}{\bar\alpha_t}}\,\epsilon'.
\]

\noindent\textbf{1) Expand $x_{t+\tau}$ in terms of $(x_0,\epsilon_t,\epsilon')$:}
\[
\begin{aligned}
x_{t+\tau}
&=\sqrt{\frac{\bar\alpha_{t+\tau}}{\bar\alpha_t}}
\Big(\sqrt{\bar\alpha_t}\,x_0+\sqrt{1-\bar\alpha_t}\,\epsilon_t\Big)
+\sqrt{1-\frac{\bar\alpha_{t+\tau}}{\bar\alpha_t}}\,\epsilon'\\
&=\sqrt{\bar\alpha_{t+\tau}}\,x_0
+\sqrt{\frac{\bar\alpha_{t+\tau}}{\bar\alpha_t}}\,\sqrt{1-\bar\alpha_t}\,\epsilon_t
+\sqrt{1-\frac{\bar\alpha_{t+\tau}}{\bar\alpha_t}}\,\epsilon'.
\end{aligned}
\]

\noindent\textbf{2) Cross-covariance $\mathrm{Cov}(x_t,x_{t+\tau})$:}
Using independence and $\mathrm{Cov}(x_0)=\Sigma,\ \mathrm{Cov}(\epsilon_t)=I,\ \mathrm{Cov}(\epsilon')=I$,
\[
\begin{aligned}
\mathrm{Cov}(x_t,x_{t+\tau})
&=\mathrm{Cov}\Big(\sqrt{\bar\alpha_t}\,x_0+\sqrt{1-\bar\alpha_t}\,\epsilon_t,\ 
\sqrt{\bar\alpha_{t+\tau}}\,x_0
+\sqrt{\frac{\bar\alpha_{t+\tau}}{\bar\alpha_t}}\,\sqrt{1-\bar\alpha_t}\,\epsilon_t
+\sqrt{1-\frac{\bar\alpha_{t+\tau}}{\bar\alpha_t}}\,\epsilon'\Big)\\
&=\sqrt{\bar\alpha_t\bar\alpha_{t+\tau}}\ \mathrm{Cov}(x_0,x_0)
+\sqrt{1-\bar\alpha_t}\ \sqrt{\frac{\bar\alpha_{t+\tau}}{\bar\alpha_t}}\,\sqrt{1-\bar\alpha_t}\ \mathrm{Cov}(\epsilon_t,\epsilon_t)\\
&=\sqrt{\bar\alpha_t\bar\alpha_{t+\tau}}\ \Sigma
+\sqrt{\frac{\bar\alpha_{t+\tau}}{\bar\alpha_t}}\,(1-\bar\alpha_t)\ I.
\end{aligned}
\]
Therefore, componentwise,
\[
\mathrm{Cov}\!\left(x_t^{(i)},x_{t+\tau}^{(j)}\right)
=\sqrt{\bar\alpha_t\bar\alpha_{t+\tau}}\ \Sigma_{ij}
+\sqrt{\frac{\bar\alpha_{t+\tau}}{\bar\alpha_t}}\,(1-\bar\alpha_t)\ \delta_{ij}.
\]

\noindent\textbf{3) Marginal variances at each time:}
\[
\begin{aligned}
\mathrm{Var}\!\left(x_t^{(i)}\right)
&=\mathrm{Var}\!\left(\sqrt{\bar\alpha_t}\,x_0^{(i)}+\sqrt{1-\bar\alpha_t}\,\epsilon_t^{(i)}\right)
=\bar\alpha_t\,\Sigma_{ii}+(1-\bar\alpha_t),\\[4pt]
\mathrm{Var}\!\left(x_{t+\tau}^{(j)}\right)
&=\mathrm{Var}\!\left(\sqrt{\bar\alpha_{t+\tau}}\,x_0^{(j)}+\sqrt{1-\bar\alpha_{t+\tau}}\,\tilde\epsilon^{(j)}\right)
=\bar\alpha_{t+\tau}\,\Sigma_{jj}+(1-\bar\alpha_{t+\tau}),
\end{aligned}
\]
(where $\tilde\epsilon$ is standard normal noise independent of $x_0$.)

\noindent\textbf{4) Correlation:}
\[
\begin{aligned}
\mathrm{Corr}\!\left(x_t^{(i)},x_{t+\tau}^{(j)}\right)
&=
\frac{\mathrm{Cov}\!\left(x_t^{(i)},x_{t+\tau}^{(j)}\right)}
{\sqrt{\mathrm{Var}(x_t^{(i)})\,\mathrm{Var}(x_{t+\tau}^{(j)})}}\\[6pt]
&=
\frac{
\sqrt{\bar\alpha_{t+\tau}\bar\alpha_t}\ \Sigma_{ij}
+\sqrt{\frac{\bar\alpha_{t+\tau}}{\bar\alpha_t}}\,(1-\bar\alpha_t)\ \delta_{ij}
}{
\sqrt{\big(\bar\alpha_t \Sigma_{ii}+(1-\bar\alpha_t)\big)\big(\bar\alpha_{t+\tau}\Sigma_{jj}+(1-\bar\alpha_{t+\tau})\big)}
}.
\end{aligned}
\]

\end{CJK*}
\end{document}